\def\year{2022}\relax
\documentclass[letterpaper]{article} 
\usepackage{aaai22}  
\usepackage{times}  
\usepackage{helvet}  
\usepackage{courier}  
\usepackage[hyphens]{url}  
\usepackage{graphicx} 
\usepackage{natbib}  
\usepackage{caption} 
\DeclareCaptionStyle{ruled}{labelfont=normalfont,labelsep=colon,strut=off} 
\usepackage{algorithm}
\usepackage{algorithmic}

\usepackage{newfloat}
\usepackage{listings}
\floatstyle{ruled}
\newfloat{listing}{tb}{lst}{}
\floatname{listing}{Listing}
\usepackage{multirow}
\usepackage[utf8]{inputenc} 
\usepackage[T1]{fontenc}    
\usepackage{hyperref}       
\usepackage{url}            
\usepackage{booktabs}       
\usepackage{amsfonts}       
\usepackage{nicefrac}       
\usepackage{microtype}      
\usepackage{amssymb,amsmath,amsfonts}
\usepackage{color}
\usepackage{graphicx}
\usepackage{xspace,soul}
\usepackage{epsfig,subfigure}
\usepackage{amsthm}
\usepackage{soul}
\usepackage{ifpdf}

\newtheorem{lemma}{Lemma}

\def\0{{\mathbf 0}}
\def\1{{\mathbf 1}}

\def\b{{\mathbf b}}
\def\c{{\mathbf c}}

\def\e{{\mathbf e}}

\def\g{{\mathbf g}}

\def\v{{\mathbf v}}

\def\x{{\mathbf x}}
\def\y{{\mathbf y}}
\def\z{{\mathbf z}}

\def\A{{\mathbf A}}
\def\B{{\mathbf B}}

\def\D{{\mathbf D}}

\def\H{{\mathbf H}}

\def\L{{\mathbf L}}
\def\M{{\mathbf M}}

\def\P{{\mathbf P}}
\def\Q{{\mathbf Q}}

\def\S{{\mathbf S}}

\def\W{{\mathbf W}}
\def\X{{\mathbf X}}

\def\ie{{\textit{i.e.}}}
\def\eg{{\textit{e.g.}}}

\def\cE{{\mathcal E}}

\def\cG{{\mathcal G}}

\def\cL{{\mathcal L}}

\def\cO{{\mathcal O}}

\def\cS{{\mathcal S}}

\def\cV{{\mathcal V}}

\def\balpha{{\boldsymbol \alpha}}

\title{Projection-free Graph-based Classifier Learning using \\ Gershgorin Disc Perfect Alignment}

\author {
    Cheng Yang,\textsuperscript{\rm 1}
    Gene Cheung, \textsuperscript{\rm 2}
    Guangtao Zhai \textsuperscript{\rm 3}
}
\affiliations {
    \textsuperscript{\rm 1} Shanghai University of Electric Power, Shanghai, China\\
    \textsuperscript{\rm 2} York University, Toronto, Canada\\
    \textsuperscript{\rm 3} Shanghai Jiao Tong University, Shanghai, China\\
    cheng.yang@shiep.edu.cn, genec@yorku.ca, zhaiguangtao@sjtu.edu.cn
}

\begin{document}

\maketitle

\begin{abstract}
In semi-supervised graph-based binary classifier learning, a subset of known labels $\hat{x}_i$ are used to infer unknown labels, assuming that the label signal $\x$ is smooth with respect to a similarity graph specified by a Laplacian matrix.
When restricting labels $x_i$ to binary values, the problem is NP-hard.
While a conventional semi-definite programming relaxation (SDR) can be solved in polynomial time using, for example, the alternating direction method of multipliers (ADMM), the complexity of projecting a candidate matrix $\M$ onto the positive semi-definite (PSD) cone ($\M \succeq 0$) per iteration remains high.
In this paper, leveraging a recent linear algebraic theory called Gershgorin disc perfect alignment (GDPA), we propose a fast projection-free method by solving a sequence of linear programs (LP) instead.
Specifically, we first recast the SDR to its dual, where a feasible solution $\H \succeq 0$ is interpreted as a Laplacian matrix corresponding to a balanced signed graph minus the last node.
To achieve graph balance, we split the last node into two, each retains the original positive / negative edges, resulting in a new Laplacian $\bar{\H}$.
We repose the SDR dual for solution $\bar{\H}$, then replace the PSD cone constraint $\bar{\H} \succeq 0$ with linear constraints derived from GDPA---sufficient conditions to ensure $\bar{\H}$ is PSD---so that the optimization becomes an LP per iteration.
Finally, we extract predicted labels from converged solution $\bar{\H}$. 
Experiments show that our algorithm enjoyed a $28\times$ speedup over the next fastest scheme while achieving comparable label prediction performance. 
\end{abstract}

\section{Introduction}
\label{sec:intro}
Binary classification---assignment of labels $\x \in \{-1, 1\}^N$ to an $N$-sample set to separate two distinct classes---is a basic machine learning problem \cite{10.5555/1162264}. 
One common setting is semi-supervised graph classifier learning: use $M$ known labels, $\hat{x}_i, 1 \leq i \leq M$, to infer $N-M$ unknown labels $x_i$, $M+1 \leq i \leq N$, assuming that $\x$ is smooth with respect to (w.r.t.) a similarity graph $\cG$ specified by a graph Laplacian matrix $\L$ \cite{zhou03,belkin04,guillory09}.
This binary graph classifier problem is NP-hard \cite{5447068}.

\textit{Semi-definite programming} (SDP) relaxation (SDR) \cite{li08} is known to provide good error-bounded approximations to \textit{quadratically constrained quadratic programs} (QCQP), of which binary graph classification is a special case (See Table I and II in \cite{5447068}). 
In a nutshell, SDR replaces the binary label constraint with a more relaxed \textit{positive semi-definite} (PSD) cone constraint (\ie, matrix variable $\M$ related to $\x \x^{\top}$ satisfying $\M \succeq 0$). 
The relaxed problem can be solved in polynomial time using, for example, the \textit{alternating direction method of multipliers} (ADMM) \cite{odonoghue16}.
However, ADMM still requires projection to the PSD cone $\cS = \{\M \,|\, \M \succeq 0\}$ per iteration, which is expensive ($\cO(N^3)$) due to full-matrix 
eigen-decomposition. 
An alternative approach removes the binary constraint and minimizes directly a quadratic graph smoothness term called \textit{graph Laplacian regularization} (GLR) $\x^\top \L \x$ \cite{pang17} for $\x \in \mathbb{R}^N$, then rounds $x_i$'s to nearest binary values $\{-1, 1\}$.
However, spectral methods such as GLR do not have tight performance bounds common in SDR \cite{goemans95}.

\begin{figure}[t]
\begin{small}
\begin{align*}
\hspace{-0.0in}\mathbf{M}=\hspace{-0.06in}\left[ \begin{array}{ccc}
2 & -2 & -1\\
-2 & 5 & -2\\
-1 & -2 & 4\\
\end{array} \right]
\hspace{-0.0in}\mathbf{SMS}^{-1}\hspace{-0.05in}=\hspace{-0.06in}
\left[ \begin{array}{ccc}
2 & -1.30 & -0.59\\
-3.07 & 5 & -1.82\\
-1.69 & -2.20 & 4\\
\end{array} \right]
\end{align*} 
\end{small}
\vspace{-0.3in}
\begin{center}
\hspace{0.1in}
\includegraphics[width=0.99\linewidth]{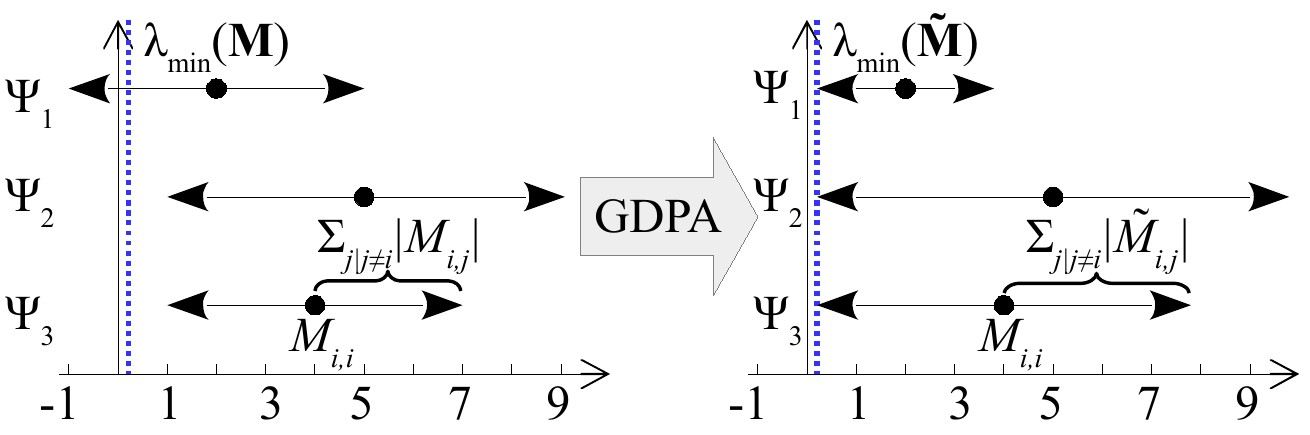}
\end{center}
\vspace{-0.15in}
\caption{\small Example of a PD matrix $\M$ and its similarity transform $\tilde{\M} = \S \M \S^{-1}$, and their respective Gershgorin discs $\Psi_i$. 
Note that Gershgorin disc left-ends of $\tilde{\M}$ are aligned at $\lambda_{\min}(\M) = 0.1078$.}
\label{fig:GDPA}
\end{figure}

To ensure matrix variable $\M$ is PSD without eigen-decomposition, one na\"{i}ve approach is to enforce linear constraints derived directly from the \textit{Gershgorin circle theorem} (GCT) \cite{gct}.
By GCT, every real eigenvalue $\lambda$ of a real symmetric matrix $\M$ resides inside at least one \textit{Gershgorin disc} $\Psi_i$---corresponding to row $i$ of $\M$---with center $c_i(\M) \triangleq M_{i,i}$ and radius $r_i(\M) \triangleq \sum_{j \neq i} |M_{i,j}|$, \ie,
\begin{align}
c_i(\M) - r_i(\M) \leq \lambda \leq c_i(\M) + r_i(\M), ~~~ \exists i.
\end{align}
The corollary is that the smallest eigenvalue, $\lambda_{\min}(\M)$, of $\M$ is lower-bounded by the smallest Gershgorin disc left-end, denoted by $\lambda^-_{\min}(\M)$, \ie,
\begin{align}
\lambda^-_{\min}(\M) \triangleq \min_i c_i(\M) - r_i(\M) \leq \lambda_{\min}(\M) .
\label{eq:GCT_bound}
\end{align}
Thus, to ensure $\M \succeq 0$, one can impose the sufficient condition $\lambda^-_{\min}(\M) \geq 0$. 
While replacing the PSD cone constraint with a set of $N$ linear constraints, $c_i(\M) - r_i(\M) \geq 0, \forall i$, is attractive computationally, GCT lower bound $\lambda^-_{\min}(\M)$ tends to be loose.
As an example, consider the \textit{positive definite} (PD) matrix $\M$ in Fig.\;\ref{fig:GDPA}(a) with $\lambda_{\min}(\M) = 0.1078$. 
The first Gershgorin disc left-end is $c_1(\M) - r_1(\M) = 2 - 3 = -1$, and $\lambda^-_{\min}(\M) < 0$.
Thus, imposing $\lambda^-_{\min}(\M) \geq 0$ directly would aggressively restrict the search space, resulting in a sub-optimal solution to the posed problem.

A recent linear algebraic theory called \textit{Gershgorin disc perfect alignment} (GDPA) \cite{yang2021signed} provides a theoretical foundation to tighten the GCT lower bound.
Specifically, GDPA states that given a graph Laplacian matrix $\L$ corresponding to a \textit{balanced} signed graph $\cG$ \cite{cartwright56}, one can perform a \textit{similarity transform}\footnote{A similarity transform $\B = \S \A \S^{-1}$ and the original matrix $\A$ share the same set of eigenvalues \cite{gct}.}, 
$\tilde{\L} = \S \L \S^{-1}$, where $\S = \text{diag}(v_1^{-1}, \ldots, v_N^{-1})$ and $\v$ is the first eigenvector of $\L$, 
such that all Gershgorin disc left-ends of $\tilde{\L}$ are \textit{exactly} aligned at $\lambda_{\min}(\L) = \lambda_{\min}(\tilde{\L})$. 
Thus, transformed $\tilde{\L}$ satisfies $\lambda^-_{\min}(\tilde{\L}) = \lambda_{\min}(\tilde{\L})$; \ie, \textit{the GCT lower bound is the tightest possible after an appropriate similarity transform}. 
Continuing our example, similarity transform $\tilde{\M} = \S \M \S^{-1}$ has all its disc left-ends exactly aligned at $\lambda_{\min}(\M) = \lambda_{\min}(\tilde{\M}) = 0.1078$. 

Leveraging GDPA, we develop a fast projection-free algorithm for semi-supervised graph classifier learning.
We first derive an SDR formulation for matrix solution $\M$ from the original graph classifier problem. 
However, solution $\M$ is not a Laplacian to a balanced graph, as required by GDPA.  
Thus, we convert the problem to its SDR dual \cite{gartner12} and interpret the dual variable $\H$ as a Laplacian to a balanced graph \textit{minus} the last graph node.
To achieve graph balance, we split the last node into two and divide the original positive and negative edges between them, resulting in a revised Laplacian $\bar{\H}$.
We repose the SDR dual problem for solution $\bar{\H}$, then replace the PSD cone constraint $\bar{\H} \succeq 0$ with linear constraints derived from GDPA.
This changes the optimization to a \textit{linear program} (LP) per iteration, which is solved efficiently
using a fast LP solver \cite{vanderbei21}. 
Finally, we extract prediction labels from converged solution $\bar{\H}$. 
Experiments show that our algorithm enjoyed $28 \times$ speedup on average over the next fastest scheme,  while retaining comparable label prediction performance.

\section{Related Work}
\label{sec:related}
Graph-based classification was first studied two decades ago \cite{zhou03,belkin04,guillory09}. With the advent of \textit{graph signal processing} (GSP) \cite{ortega18ieee,cheung18}---analysis of discrete signals residing on finite graphs---interest in the problem was revived \cite{gavish10,shuman11,cheung18tsipn}. 
The problem of learning a similarity graph from data has been extensively studied \cite{dong19}. 
We focus on the orthogonal problem of predicting binary labels given a graph and a subset of $M$ labels.

SDR---useful in approximating NP-hard problems \cite{gartner12} such as QCQP---provides an effective relaxation to the binary graph classifier problem \cite{li08}.
An interior point method tailored for the slightly more general \textit{binary quadratic problem}\footnote{BQP objective takes a quadratic form $\x^{\top} \Q \x$, but $\Q$ is not required to be a Laplacian to a similarity graph.} (BQP) has complexity $\cO(N^{3.5} \log (1/\epsilon))$, where $\epsilon$ is the tolerable error \cite{helmberg96}.
The complexity was improved to $\cO(N^3)$ by SDCut
\cite{wang13,7431988} via spectrahedron-based relaxation.
Replacing the PSD cone constraint $\M \succeq 0$ with a factorization $\M = \X \X^{\top}$ was proposed in \cite{shah16}, but resulted in a non-convex optimization for $\X$ that was minimized locally, where in each iteration a matrix inverse of worst-case complexity $\cO(N^3)$ was required.  
More recent first-order methods for SDP such as \cite{odonoghue16} used ADMM \cite{boyd11,8571259,cdcs_2020},
but the iterative projection onto PSD cone requires full-matrix eigen-decomposition and thus expensive.
In contrast, leveraging GDPA theory \cite{yang2021signed}, our algorithm is entirely projection-free.

It is known in graph spectral theory \cite{chung96} that balanced signed graphs have unique spectral properties \cite{dittrich20}; for example, the \textit{signed graph Laplacian matrix} \cite{kunegis10} has eigenvalue $0$ iff the underlying graph is balanced. 
In contrast, 
GDPA \cite{yang2021signed} states that all Gershgorin disc left-ends of a similarity transform $\S \M \S^{-1}$ of graph Laplacian $\M$ to a balanced graph can be perfectly aligned at $\lambda_{\min}(\M)$. 
GDPA theory was developed for fast \textit{metric learning} \cite{moutafis17} to optimize a PD matrix $\M$ given a convex and differentiable objective $Q(\M)$.

While also leveraging GDPA, this work addresses the binary graph classification problem in a different and non-trivial manner. 
Specifically, observing that solution $\H$ to the SDR dual is a Laplacian to a balanced graph $\cG$ minus the last node, we augment the last node to obtain an overall balanced graph $\bar{\cG}$ via new Lemma\;\ref{lemma:Hbar}, and solve a modified SDR dual for Laplacian $\bar{\H}$ to $\bar{\cG}$ via GDPA linearization.

\section{Preliminaries}
\label{sec:prelim}
\subsection{Graph Definitions}

A graph $\cG(\cV,\cE,\W)$ has node set $\cV = \{1 \ldots, N\}$ and edge set  $\cE = \{ (i,j)\}$, where $(i,j)$ means nodes $i$ and $j$ are connected with edge weight $w_{i,j} \in \mathbb{R}$.
A \textit{positive} graph means $w_{i,j} \geq 0, \forall (i,j) \in \cE$, while a \textit{signed} graph means $w_{i,j}$ can be negative as well.
A node $i$ may have self-loop of weight $u_i \in \mathbb{R}$. 
Denote by $\W$ the \textit{adjacency matrix}, where $W_{i,j} = w_{i,j}$ if $(i,j) \in \cE$ and $= 0$ otherwise, and $W_{i,i} = u_i$. 
We assume undirected edges, and thus $\W$ is symmetric.
Define the diagonal \textit{degree matrix} $\D$, where $D_{i,i} = d_i \triangleq \sum_{j} W_{i,j}$ is the degree of node $i$. 
The \textit{combinatorial graph Laplacian matrix} \cite{ortega18ieee} is defined as $\L \triangleq \D - \W$.
To account for self-loops, the \textit{generalized graph Laplacian matrix} is $\cL \triangleq \D - \W + \text{diag}(\W)$.
Note that any real symmetric matrix can be interpreted as a generalized graph Laplacian matrix. 

The \textit{graph Laplacian regularizer} (GLR) \cite{pang17} that quantifies smoothness of signal $\x \in \mathbb{R}^N$ w.r.t. graph specified by $\cL$ is
\begin{align}
\x^{\top} \cL \x = \sum_{(i,j) \in \cE} w_{i,j} (x_i - x_j)^2 + \sum_{i \in \cV} u_i x_i^2 .
\label{eq:glr}
\end{align}
GLR is also the objective of our graph classification problem.

\subsection{Iterative GDPA Linearization} 
Denote by $\cL$ a generalized graph Laplacian matrix to a balanced and connected signed graph $\cG$ (with or without self-loops).
A \textit{balanced} graph has no cycle of odd number of negative edges.
By the \textit{Cartwright-Harary Theorem} (CHT) \cite{cartwright56}, a graph is balanced iff nodes can be colored into blue and red, such that positive (negative) edges connect nodes of the same (different) colors.

GDPA \cite{yang2021signed} states that
a similarity transform $\tilde{\cL} = \S \cL \S^{-1}$, where $\S = \text{diag}(s_1, \ldots, s_N)$, $s_i = v_i^{-1}, \forall i$, and $\v$ is the provably strictly non-zero first eigenvector  of $\cL$, has all its Gershgorin disc left-ends aligned exactly at smallest eigenvalue $\lambda_{\min}(\cL)$, \ie,
\begin{align}
\begin{split}
\tilde{\cL}_{i,i} - \sum_{j \neq i} | \tilde{\cL}_{i,j} | & = \cL_{i,i} - \sum_{j \neq i} | s_i \cL_{i,j} / s_j | \\ & = \lambda_{\min}(\cL),
~~~~~ \forall i \in \{1, \ldots, N\}.
\end{split}
\end{align}
To solve an optimization of the form $\min_{\cL \succeq 0} Q(\cL)$, one can leverage GDPA and optimize iteratively as follows.
At iteration $t$ with previously computed solution $\cL^t$, compute first eigenvector $\v^t$ to $\cL^t$ corresponding to $\lambda_{\min}(\cL^t)$; extreme eigenvector $\v^t$ can be computed in linear-time complexity $\cO(N)$ using \textit{Locally Optimal Block Preconditioned Conjugate Gradient} (LOBPCG) \cite{Knyazev01} assuming a sparse matrix\footnote{For computation reasons, Laplacian $\cL$ is typically sparse to specify a sparse graph $\cG$ in the GSP literature \cite{ortega18ieee}.}
Define scalars $s^t_i = 1/v^t_i, \forall i$, then solve
\begin{align}
\begin{split}
\min_{\cL} &~~ Q(\cL), \\
\text{s.t.} &~~ 
\cL_{i,i} - \sum_{j \neq i} |s^t_i \cL_{i,j}/s^t_j| \geq 0, ~~~ \forall i \in \{1, \ldots, N\} .
\label{eq:GCT_linConst}
\end{split}
\end{align}
Linear constraints in \eqref{eq:GCT_linConst} ensure that the similarity transform $\tilde{\cL} = \S \cL \S^{-1}$ is PSD by GCT, and hence solution $\cL$ is PSD. 
Since scalars $\{s^t_i\}$ are computed from first eigenvector $\v^t$ of $\cL^t \succeq 0$, by GDPA, similarity transform $\S \cL^t \S^{-1}$ has all its disc left-ends aligned exactly at $\lambda_{\min}(\cL^t) \geq 0$, and hence $\cL^t$ remains feasible at iteration $t$.
Thus, objective $Q(\cL^t)$ is monotonically non-increasing with $t$, and the algorithm converges to a local minimum\footnote{See the supplement for an exposition of local convergence.}. 
We invoke this iterative procedure to solve our posed SDR dual in the sequel.

\section{Binary Graph Classification}
\label{sec:formulate}
We first formulate the binary graph classification problem and relax it to an SDR problem.
We then present its SDR dual with dual variable matrix $\H$.
Finally, we interpret $\H$ as a graph Laplacian, and augment its corresponding graph $\cG$ to a balanced graph $\bar{\cG}$ for GDPA linearization.

\subsection{SDR Primal}
\label{subsec:SDP_primal}

Given a PSD graph Laplacian matrix $\L \in \mathbb{R}^{N \times N}$ of a positive similarity graph $\cG^o$, one can formulate a binary graph classification problem as
\begin{align}
\min_{\x} \x^{\top} \L \x, 
~~~~ \mbox{s.t.}~ \left\{
\begin{array}{l}
x_i^2 = 1, ~\forall i \in \{1, \ldots, N\} \\
x_i = \hat{x}_i, ~\forall i \in \{1, \ldots, M\}
\end{array} \right. 
\label{eq:binaryClass}
\end{align}
where $\{\hat{x}_i\}_{i=1}^M$ are the $M$ known labels. 
The objective in \eqref{eq:binaryClass} dictates that signal $\x$ is smooth w.r.t. graph $\cG^o$ specified by $\L$. 
Because $\L$ is PSD \cite{cheung18}, the objective is lower-bounded by $0$, \ie, $\x^{\top} \L \x \geq 0, \forall \x \in \mathbb{R}^N$.
The first binary constraint ensures $x_i \in \{-1, 1\}$. 
The second constraint ensures that entries $x_i$ in signal $\x$ agree with known labels $\{\hat{x}_i\}_{i=1}^M$. 

\begin{figure}[t]
\begin{center}
\ifpdf
\includegraphics[width=0.95\linewidth]{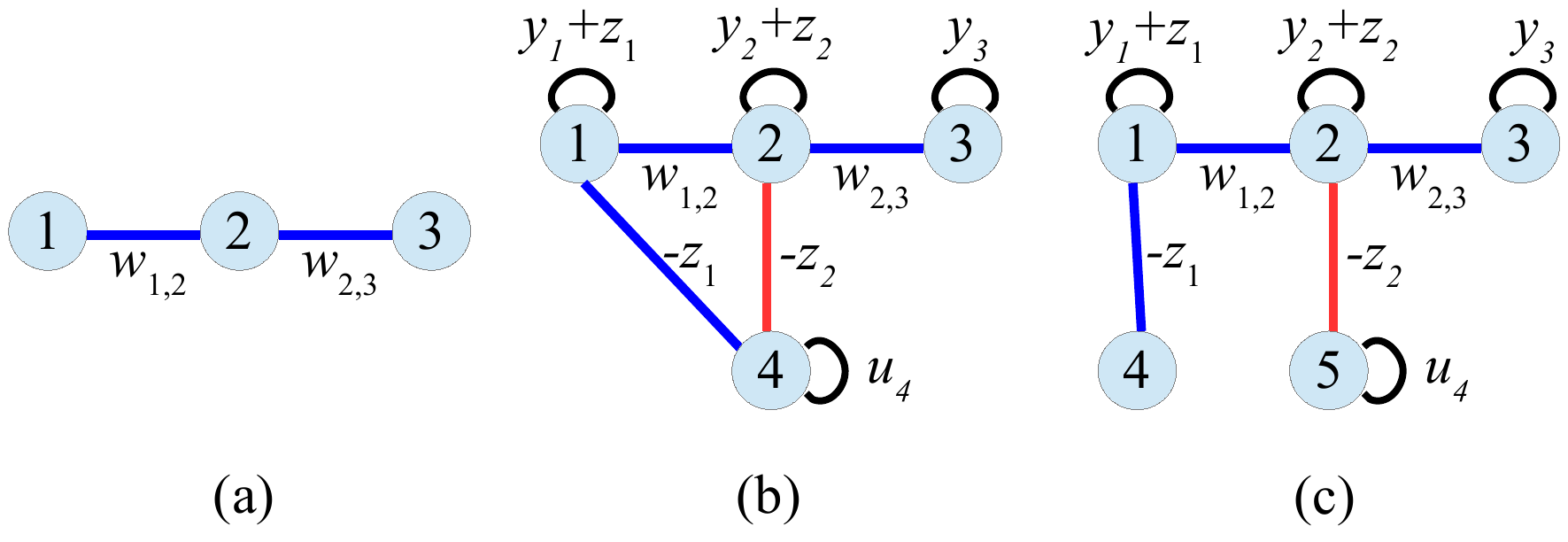}
\else
 do something for regular latex or pdflatex in dvi mode
\fi
\end{center}
\vspace{-0.15in}
\caption{(a) 3-node line graph example. 
(b) Solution $\H$ to SDR dual \eqref{eq:SDP_dual} as graph Laplacian matrix. 
(c) Solution $\bar{\H}$ to modified SDR dual \eqref{eq:SDP_dual2} as graph Laplacian matrix. Positive / negative edges are colored in blue / red. Self-loop weight $u_4$ in (b) for node $4$ is $u_4 = y_4 + z_1 + z_2$.}
\label{fig:3-node-graph}
\end{figure}

As an example, consider a 3-node line graph shown in Fig.\;\ref{fig:3-node-graph}(a), where edges $(1,2)$ and $(2,3)$ have weights $w_{1,2}$ and $w_{2,3}$, respectively. 
The corresponding adjacency matrix $\W$ and graph Laplacian matrix $\L$ are
\begin{align}
\begin{split}
\W &= \left[ \begin{array}{ccc}
0 & w_{1,2} & 0 \\
w_{1,2} & 0 & w_{2,3} \\
0 & w_{2,3} & 0
\end{array} \right],\\  
\L &= \left[ \begin{array}{ccc}
w_{1,2} & -w_{1,2} & 0 \\
-w_{1,2} & w_{1,2} + w_{2,3} & -w_{2,3} \\
0 & -w_{2,3} & w_{2,3}
\end{array} \right] .
\end{split} 
\end{align}
Suppose known labels are $\hat{x}_1 = 1$ and $\hat{x}_2 = -1$. 

\eqref{eq:binaryClass} is NP-hard \cite{5447068}. 
One can derive a corresponding SDR \cite{5447068} as follows.
Define first $\X = \x \x^{\top}$ and $\M = [\X ~~ \x; ~\x^{\top} ~~ 1]$.
$\M$ is PSD because: 
i) sub-block $1$ is trivially PSD, and ii) the \textit{Schur complement} of sub-block $1$ of $\M$ is $\X - \x \x^{\top} = \0$, which is also PSD.
Thus, $\X = \x \x^{\top}$ (or equivalently $\text{rank}(\X) = 1$) implies $\M \succeq 0$, but not vice versa.
$\X = \x \x^{\top}$ and $X_{ii} = 1, \forall i$ imply $x_i^2 = 1, \forall i$. 
To convexify the problem, we relax $\X = \x \x^{\top}$ to $\M \succeq 0$ and write the SDR for optimization variable $\M$ as
\begin{align}
\min_{\x,\X} \text{Tr}(\L \X) ~\mbox{s.t.}~
\left\{ \begin{array}{l}
\M \triangleq \left[ \begin{array}{cc}
\X & \x \\
\x^{\top} & 1
\end{array} \right] \succeq 0 \\
X_{ii} = 1, ~ i \in \{M+1, \ldots, N\} \\
x_i = \hat{x}_i, ~ i \in \{1, \ldots, M\}
\end{array} \right. 
\label{eq:primal}
\end{align}
\noindent
where $\text{Tr}(\x^{\top} \L \x) = \text{Tr}(\L \x \x^{\top}) = \text{Tr}(\L \X)$. 
Because \eqref{eq:primal} has linear objective and constraints with an additional PSD cone constraint, $\M \succeq 0$, it is an SDP problem \cite{gartner12}. 
We call \eqref{eq:primal} the \textit{SDR primal}.

Continuing our example, consider ground-truth labels $\x = [1 ~~-1 ~~1]^{\top}$ for the 3-node graph in Fig.\;\ref{fig:3-node-graph}(a).  
The corresponding solution matrix $\M = [\x \x^{\top} ~\x; ~\x^{\top} ~1]$ is
\begin{align}
\M = \left[ \begin{array}{cccc}
1 & -1 & 1 & 1\\
-1 & 1 & -1 & -1 \\
1 & -1 & 1 & 1 \\
1 & -1 & 1 & 1
\end{array} \right] .
\end{align}
Observe that $\M$ is not a graph Laplacian matrix corresponding to a balanced signed graph, as required by GDPA. 
This motivates us to investigate the corresponding SDP dual.

\subsection{SDR Dual}
\label{subsec:SDP_dual}

We write the corresponding dual problem based on SDP duality theory \cite{gartner12}.
We first define 
\begin{align}
\A_i = \text{diag}(\e_{N+1}(i)),~~~~
\B_i = \left[ \begin{array}{cc}
\0_{N \times N} & \e_N(i) \\
\e^{\top}_N(i) & 0
\end{array} \right] 
\label{eq:AB}
\end{align}
where $\e_N(i) \in \{0, 1\}^{N}$ is a length-$N$ binary \textit{canonical vector} with a single non-zero entry equals to $1$ at the $i$-th entry, $\0_{N \times N}$ is a $N$-by-$N$ matrix of zeros, and $\text{diag}(\v)$ is a diagonal matrix with diagonal entries equal to $\v$.
Note that both $\A_i$ and $\B_i$ are symmetric. 

Next, we collect $M$ known labels $\{\hat{x}_i\}_{i=1}^M$ into a vector $\b \in \mathbb{R}^{M}$ of length $M$, \textit{i.e.},
\begin{align}
b_i &= 2 \hat{x}_i, ~~~~~~~~ \forall i \in \{1, \ldots, M\} .
\label{eq:bc}
\end{align}
We now define the SDR dual of \eqref{eq:primal} as
\begin{align}
\min_{\y, \z} & ~~ \1^{\top}_{N+1} \y + \b^{\top} \z, 
\label{eq:SDP_dual} \\
\mbox{s.t.} & ~~
\H \triangleq \sum_{i=1}^{N+1} y_i \A_i +
\sum_{i=1}^M z_i \B_i + \left[ \begin{array}{cc}
\L & \0_N \\
\0_N^\top & 0 
\end{array} \right] \succeq 0
\nonumber 
\end{align} \noindent
where $\1_N$ is a length-$N$ vector of ones, and dual variables are $\y\in\mathbb{R}^{N+1}$ and $\z\in\mathbb{R}^M$.
Given that \eqref{eq:SDP_dual} is a minimization, when $b_i < 0$ (\ie, $\hat{x}_i < 0$), the corresponding $z_i$ is non-negative\footnote{$z_i < 0$ when $b_i < 0$ would mean a worse objective \textit{and}  larger Gershgorin disc radii for rows $i$ and $N+1$ of matrix $\H$, making $\H$ more difficult to reside in the PSD cone $\H \succeq 0$.}, \ie, $z_i \geq 0$.
Similarly, for $b_i > 0$, $z_i \leq 0$.
Thus, the signs of $z_i$'s are known \textit{a priori}.
Without loss of generality, we assume $z_i \leq 0, \forall i \in \{1, \ldots, M_1\}$ and $z_i \geq 0, \forall i \in \{M_1+1, \ldots, M\}$, where $1 \leq M_1 < M$, in the sequel.

\subsection{Reformulating the SDR Dual}
\label{subsec:reformulate}

We interpret $\H \in \mathbb{R}^{(N+1) \times (N+1)}$ in \eqref{eq:SDP_dual} as a graph Laplacian corresponding to a signed graph $\cG$.
However, \textit{$\cG$ is not balanced}, because of the last row / column in $\H$.
To see this, we write
\begin{align}
\H = \left[ \begin{array}{cc}
\cL_y & \g \\
\g^{\top} & y_{N+1}
\end{array} \right]
\end{align}
where $\g = [z_1 \ldots z_M ~ \0_{N-M}^{\top}]^{\top}$.
Matrix $\cL_y \in \mathbb{R}^{N \times N}$, defined as $\cL_y \triangleq \text{diag}(y_1, \ldots, y_N) + \L$, is a generalized Laplacian to a $N$-node positive graph $\cG^+$.
However, node $N+1$ has \textit{both} positive \textit{and} negative edges to $\cG^+$ stemming from negative $z_i$'s and positive $z_i$'s, respectively. 
As a result, $\H$ is not a Laplacian to a balanced signed graph.

Continuing our 3-node line graph example with Laplacian $\L$, the corresponding $\cL_y$ and $\H$ are

\vspace{-0.05in}
\begin{small}
\begin{align}
\begin{split}
\cL_y &= \left[ \begin{array}{ccc}
y_1 + w_{1,2} & -w_{1,2} & 0 \\
-w_{1,2} & y_2 + w_{1,2} + w_{2,3} & -w_{2,3} \\
0 & -w_{2,3} & y_3 + w_{2,3}
\end{array} \right],\\
\H &= \left[ \begin{array}{cccc}
y_1 + w_{1,2} & -w_{1,2} & 0 & z_1 \\
-w_{1,2} & y_2 + w_{1,2} + w_{2,3} & -w_{2,3} & z_2 \\
0 & -w_{2,3} & y_3 + w_{2,3} & 0 \\
z_1 & z_2 & 0 & y_4
\end{array} \right] .
\end{split}
\label{eq:ex_H}
\end{align}
\end{small}\noindent
Interpreting $\H$ as a graph Laplacian, node $1$ has degree $d_1 = y_1 + w_{1,2} = u_1 + w_{1,2} - z_1$, and thus $u_1 = y_1 + z_1$. Similarly, node $4$ has degree $d_4 = y_4 = u_4 - z_1 - z_2$, and thus $u_4 = y_4 + z_1 + z_2$.
See Fig.\;\ref{fig:3-node-graph}(b) for an illustration of this unbalanced signed graph $\cG$.
More generally, self-loop weights are $u_i = y_i + z_i$ for $i \in \{1, \ldots, M\}$, $u_i = y_i$ for $i \in \{M+1, \ldots, N\}$, and $u_{N+1} = y_{N+1} + \sum_{j=1}^M z_j$. 

Node $N+1$ has positive and negative edges, with respective weights $\{-z_i\}_{i=1}^{M_1}$ and $\{-z_i\}_{i={M_1+1}}^{M}$ to $\cG^+$, and a self-loop with weight $u_{N+1}$.  
We construct an \textit{augmented} graph $\bar{\cG}$ with $N+2$ nodes from $\cG$ by splitting node $N+1$ in $\cG$ into two in $\bar{\cG}$, assigning positive and negative edges to the two respectively.
The graph construction procedure is
\begin{enumerate}
\item Construct each $i$ of first $N$ nodes with the same inter-node edges as $\cG^+$ plus self-loop with weight $u_i$.
\item Construct node $N+1$ with positive edges $\{-z_i\}_{i=1}^{M_1}$, and node $N+2$ with negative edges $\{-z_i\}_{i=M_1+1}^M$, to the first $N$ nodes in sub-graph $\cG^+$.
\item Add self-loop for nodes $N+2$ with weight $u_{N+1}$.  
\end{enumerate}
Step 3 implies that nodes $N+1$ and $N+2$ have degrees $\kappa_{N+1} \triangleq - \sum_{i=1}^{M_1} z_i$ and $\kappa_{N+2} \triangleq u_{N+1} - \sum_{i=M_1+1}^M z_i$, respectively.
Denote by $\bar{\H} \in \mathbb{R}^{(N+2) \times (N+2)}$ the graph Laplacian matrix corresponding to $\bar{\cG}$. 
Continuing our 3-node graph example, Fig.\;\ref{fig:3-node-graph}(c) shows the augmented graph $\bar{\cG}$, and the corresponding Laplacian $\bar{\H}$ is

\vspace{-0.05in}
\begin{small}
\begin{align}
\bar{\H} &= \left[ \begin{array}{ccccc}
d_1 & -w_{1,2} & 0 & z_1 & 0 \\
-w_{1,2} & d_2 & - w_{2,3} & 0 & z_2 \\
0 & -w_{2,3} & d_3 & 0 & 0 \\
z_1 & 0 & 0 & - z_1 & 0 \\
0 & z_2 & 0 & 0 & u_4 - z_2
\end{array} \right] 
\end{align}
\end{small}\noindent 
where $d_i = y_i + \sum_{j \neq i} w_{i,j}$ is the degree of node $i$ for $i \in \{1, \ldots, N\}$.

Crucially, $\bar{\H}$ and $\H$ are related in spectral terms by the following important lemma.

\begin{lemma}
Smallest eigenvalue $\lambda_{\min}(\bar{\H})$ of graph Laplacian $\bar{\H}$ to augmented graph $\bar{\cG}$ is a lower bound for smallest eigenvalue $\lambda_{\min}(\H)$ of Laplacian $\H$ to $\cG$, \ie, 
\begin{align}
\lambda_{\min}(\bar{\H}) \leq \lambda_{\min}(\H) .
\end{align}
\label{lemma:Hbar}
\end{lemma}\noindent 

\vspace{-0.1in}
\begin{proof}
Denote by $\cG$ the graph represented by generalized graph Laplacian $\H$, with inter-node edge weights $\{w_{i,j}\}$ and self-loop weights $\{u_i\}$. 
Denote by $\v \in \mathbb{R}^{N+1}$ the first eigenvector of $\H$ corresponding to the smallest eigenvalue $\lambda_{\min}(\H)$. 
From \eqref{eq:glr}, GLR of $\H$ computed using $\v$ is
\begin{align}
\v^{\top} \H \v 
=& \!\!\! \sum_{(i,j) \in \cE \,|\, 1 \leq i,j \leq N} \!\!\!\!\!\! w_{i,j} (v_i - v_j)^2 -
\sum_{i=1}^M z_i (v_{N+1} - v_i)^2 \nonumber\\
& + \sum_{i=1}^{N} u_i v_i^2 + u_{N+1} v_{N+1}^2 .
\end{align}
Now construct length-($N+2$) vector $\balpha \in \mathbb{R}^{N+2}$, where $\balpha = [v_1 \ldots v_N ~v_{N+1} ~v_{N+1}]^{\top}$. 
GLR of $\bar{\H}$ using $\balpha$ is

\begin{small}
\begin{align}
\balpha^{\top} \bar{\H} \balpha =& \!\! \sum_{(i,j) \in \cE \,|\, 1\leq i,j \leq N} \!\!\! w_{i,j} (v_i - v_j)^2 -
\sum_{i=1}^{M_1} z_i (v_{N+1} - v_i)^2 \nonumber\\
&- \sum_{i=M_1+1}^M z_i (v_{N+1} - v_i)^2 + \sum_{i=1}^N u_i v_i^2 + u_{N+1} v_{N+1}^2 .
\end{align}
\end{small}\noindent 
Thus, $\v^{\top} \H \v = \balpha^{\top} \bar{\H} \balpha$.
Since first eigenvector $\v$ of $\H$ minimizes its Rayleigh quotient,
\begin{align}
\lambda_{\min}(\H) = \frac{\v^{\top} \H \v}{\v^{\top} \v} \stackrel{(a)}{\geq} \frac{\balpha^{\top} \bar{\H} \balpha}{\balpha^{\top} \balpha}
\stackrel{(b)}{\geq} \lambda_{\min}(\bar{\H}) .
\end{align}
$(a)$ holds since $\v^{\top} \v \leq \balpha^{\top} \balpha$ by construction, and $(b)$ holds since $\lambda_{\min}(\bar{\H}) = \min_{\x} \frac{\x^{\top} \bar{\H} \x}{\x^{\top} \x}$.  
\end{proof}

In our experiments, we verify numerically that the bound $\lambda_{\min}(\bar{\H}) \leq \lambda_{\min}(\H)$ was tight in realistic datasets. 

Given Lemma\;\ref{lemma:Hbar}, we reformulate the SDR dual \eqref{eq:SDP_dual} by keeping the same objective but imposing the PSD cone constraint on $\bar{\H}$ instead of $\H$.
First, define $\A'_i$, $\B'_i$ and $\B''_i$ similarly to \eqref{eq:AB} but for a larger $(N+2)$-by-$(N+2)$ matrix;
\ie, $\A'_i \triangleq \text{diag}(\e_{N+2}(i))$, 

\vspace{-0.05in}
\begin{footnotesize}
\begin{align}
\B'_i \triangleq \left[ \begin{array}{cc}
\B_i & \0_{N+1} \\ \0_{N+1}^{\top} & 0 
\end{array}\right],
~
\B''_i \triangleq \left[ \begin{array}{cc}
\0_{(N+1) \times (N+1)} & \e_{N+1}(i) \\ \e^{\top}_{N+1}(i) & 0 \end{array}\right].
\end{align}
\end{footnotesize}\noindent
The reformulated SDR dual is

\vspace{-0.05in}
\begin{align}
\min_{\y, \z} &~~ \1^{\top}_{N+1} \y + \b^{\top} \z, 
\label{eq:SDP_dual2} \\
\mbox{s.t.} &~~
\bar{\H} \triangleq \sum_{i=1}^{N} y_i \A'_i +
\kappa_{N+1} \A'_{N+1} + 
\kappa_{N+2} \A'_{N+2} \nonumber\\
& + \sum_{i=1}^{M_1} z_i \B'_i 
+ \sum_{i=M_1+1}^{M} z_i \B''_i + \left[\begin{array}{cc}
\L & \0_{N \times 2} \\
\0_{2 \times N} & \0_{2 \times 2}
\end{array}\right] \succeq 0
\nonumber
\end{align}
where $\kappa_{N+1}$ and $\kappa_{N+2}$ are the degrees of nodes $N+1$ and $N+2$, respectively, defined earlier.

We can bound the difference in objective values between the optimal solutions to \eqref{eq:SDP_dual} and \eqref{eq:SDP_dual2} as follows.  
We first construct yet another modified graph $\tilde{\cG}$ from $\cG$, where weight $-z_i$ of each edge from node $N+1$ to node $i \in \{1, \ldots, M\}$ is incremented by $\phi$. 
This results in another related graph Laplacian matrix $\tilde{\H}$ for modified $\tilde{\cG}$. 
Continuing our example, the modified graph Laplacian $\tilde{\H}$ from $\H$ in (14) is
\begin{align}
\tilde{\H} = \H + 
\left[ \begin{array}{cccc}
\phi & 0 & 0 & -\phi \\
0 & \phi & 0 & -\phi \\
0 & 0 & 0 & 0 \\
-\phi & -\phi & 0 & 2\phi
\end{array} \right] .
\end{align}
Similar to Lemma 1, we claim in the following lemma that $\tilde{\H}$ and $\H$ are related in spectral terms.

\noindent\textbf{Lemma 2.} \textit{The smallest eigenvalue $\lambda_{\min}(\H)$ of graph Laplacian $\H$ to graph $\cG$ is a lower bound for $\lambda_{\min}(\tilde{\H})$ of Laplacian $\tilde{\H}$ to $\tilde{\cG}$, \ie,}
\begin{align}
\lambda_{\min}(\H) \leq \lambda_{\min}(\tilde{\H}) .    
\end{align}
\label{lemma:Htilde}

See the proof in the supplementary file.
The corollary is that $\tilde{\H} \succeq 0$ if $\H \succeq 0$, or more simply, $\tilde{\H} \succeq \H$. 
This means that minimizing the same objective in \eqref{eq:SDP_dual} but using the more relaxed constraint $\tilde{\H} \succeq 0$ instead will yield an objective value $F(\tilde{\H})$ that is no worse than $F(\H)$. 
Given $\tilde{\H} \succeq \H \succeq \bar{\H}$, we know $F(\tilde{\H}) \leq F(\H) \leq F(\bar{\H})$. 
Thus, we can bound the approximation error $|F(\bar{\H}) - F(\H)|$ between the modified SDR dual \eqref{eq:SDP_dual2} and the original SDR dual \eqref{eq:SDP_dual} as
\begin{align}
|F(\bar{\H}) - F(\H)| \leq |F(\bar{\H}) - F(\tilde{\H})| . 
\label{eq:errBound}
\end{align}

Finally, we note that minimizing objective in \eqref{eq:SDP_dual} with constraint $\tilde{\H} \geq 0$ is much easier if $\phi$ is sufficiently large such that all edges from node $N+1$ becomes positive. 
In such case, $\tilde{\cG}$ is a positive graph, and GDPA linearization can be applied. 
Thus, the error bound \eqref{eq:errBound} can be numerically computed efficiently for each instant of SDR dual \eqref{eq:SDP_dual}.

Given $\bar{\H}$ is a Laplacian to a balanced graph, we discuss using GDPA linearization to solve \eqref{eq:SDP_dual2} next.

\section{Algorithm Implementation}
\label{sec:algo}
\subsection{GDPA Linearization}
\label{ssec:linearization}

We replace the PSD cone constraint on $\bar{\H}$ in \eqref{eq:SDP_dual2} with $N+2$ linear constraints via GDPA.
Specifically, at iteration $t$, we compute first eigenvector $\v^t$ of solution $\bar{\H}^t$ using LOBPCG \cite{Knyazev01}.
We define scalars $s_i = 1/v_i^t, \forall i \in \{1, \ldots, N+2\}$.
Finally, we write $N+2$ constraints corresponding to $\lambda^-_{\min}(\S \bar{\H} \S^{-1}) \geq 0$, where $\S = \text{diag}(s_1, \ldots, s_{N+2})$, \ie,  
\begin{align}
\bar{H}_{i,i} - \sum_{j\neq i} |s_i \bar{H}_{i,j} / s_j | \geq 0, ~~\forall i \in \{1, \ldots, N+2\} .  
\label{eq:LP_dual}
\end{align}
Note that the absolute value operation can be appropriately removed for each term $s_i \bar{H}_{i,j}/s_j$, since the signs for $s_i$ and $\bar{H}_{i,j}$ are known. 
Together with linear objective in \eqref{eq:SDP_dual2}, this constitutes an LP for variables $\y$ and $\z$, solvable using an available fast LP solver \cite{vanderbei21}\footnote{
The lowest complexity of a general LP solver \cite{jiang2020faster} to date is $\mathcal{O}(N^{2.055})$.
Note that the LP field is still fast-evolving, and our proposal is not tied to a specific LP solver.}.
Compared to SDR primal \eqref{eq:primal} with a large matrix variable $\M \in \mathbb{R}^{(N+1) \times (N+1)}$, dimensions of our LP variables, $\y \in \mathbb{R}^{N+1}$ and $\z \in \mathbb{R}^M$, are much smaller.

A sequence of LPs are solved, each time with scalars $s_i$'s updated from computed solution $\bar{\H}^t$, until convergence. 
The bulk of the complexity resides in the computation of the first eigenvector $\v^t$ for each LP solution $\bar{\H}^t$. 
LOBPCG \cite{Knyazev01} is an iterative algorithm running in linear time for extreme eigenvectors of sparse matrices, which further benefits from \textit{warm start}: with a good initial guess for $\v^t$, the algorithm converges faster.
Since $\bar{\H}^t$ changes gradually during the iterations, we use previously computed eigenvector $\v^{t-1}$ of $\bar{\H}^{t-1}$ as initial guess for $\v^t$ of $\bar{\H}^t$.
Experiments show that warm start improves convergence speed significantly.

\subsection{Initialization \& Prediction Label Extraction}


Our posed LP requires an initial $\bar{\H}^0$ to compute first eigenvector $\v^0$, so that scalars $\{s_i\}_{i=1}^{N+2}$ can be defined for $N+2$ linear constraints in \eqref{eq:LP_dual}.
To initialize $\bar{\H}^0$, we set $\y^0=[\1_M^\top \; \0_{N-M}^\top \; M]^\top$ and
$\z^0= [-\hat{x}_1 \ldots -\hat{x}_M]$.
$\bar{\H}^0$ can then be computed using definition of $\bar{\H}$ in \eqref{eq:SDP_dual2}.

As similarly done in \cite{5447068}, we extract labels $\x^* = [x_{1} \ldots x_N]^{\top}$ from converged LP solution $\y^*$ and $\z^*$ as follows.
We first construct $\H^*$ using $\y^*$ and $\z^*$ using definition of $\H$ in \eqref{eq:SDP_dual}.
We then compute $\x^*=\text{sign}(\hat{x}_1v_1\v)$, where $v_1$ is the first entry of the first eigenvector $\v$ of $\H^*$.
See \cite{5447068} for details of recovering SDP primal variables from dual variables in BQP.
Finally, we extract the prediction labels as $\tilde{\x}=[x_{M+1}^*,...,x_{N}^*]^\top$.

\section{Experiments}
\label{sec:results}

\subsection{Experimental Setup}
\label{ssec:experimentalsetup}

We implemented our GDPA classifier in Matlab\footnote{available at \url{https://anonymous.4open.science/r/gc\_-80C0}}, 
and evaluated it in terms of average classification error rate and running time.
All computations were carried out
on a Windows 10 64bit PC with AMD RyzenThreadripper
3960X 24-core processor 3.80 GHz and 128GB of RAM.
We compared our algorithm against the following schemes that solve the SDR primal problem \eqref{eq:primal} directly: 
i) two primal-dual interior-point solvers for SDP, SeDuMi and MOSEK \cite{cvx_link}, 
ii) an ADMM first-order operator-splitting solver CDCS
\cite{8571259,cdcs_link},
iii) a spectrahedron-based relaxation solver SDCut
\cite{wang13,sdcut_link} that involves L-BFGS-B \cite{10.1145/279232.279236}, and
iv) a biconvex relaxation solver BCR \cite{shah16,bcr_link}, 
all of which are implemented in Matlab. 
Further, we employed CDCS again to solve our modified SDR dual problem \eqref{eq:SDP_dual2}.
We focus our comparison with SDR schemes because, again, SDR is known to provide good error-bounded approximations in general for NP-hard QCQP problems \cite{5447068}. 

In addition, we compared against the following non-SDR methods approximating original classifier formulation \eqref{eq:binaryClass} directly. 
A recent method called \textit{stochastic neighborhood search} (SNS) \cite{9145635,sns_link} solves \eqref{eq:binaryClass} by alternately applying Karush–Kuhn–Tucker optimality condition guided deterministic search and bootstrapping
sampling based stochastic search.
We solved a relaxed version of \eqref{eq:binaryClass} using SeDuMi, where constraint $x_i^2=1$ was relaxed to a box constraint $x_i\in[-1,1]$---we denote this method by GLR-box. 
Finally, binary constraint $x_i^2=1$ can be ignored entirely, and objective $\x^\top \L \x$ in \eqref{eq:binaryClass} can be optimized simply by computing extreme eigenvectors plus rounding. We denote this class of spectral methods by SPEC, which are fast but are known to have poor worst-case errors \cite{doi:10.1137/S0895479896312262}.

\subsection{Experimental Results}
\label{ssec:results}



We first show in Table\;\ref{tab:restore} that SPEC has by far the worst performance in binary signal restoration compared to SDR-based schemes and SNS, demonstrating the limitations of spectral methods in general.
Specifically, following an illustrative example in \cite{9145635}, we first corrupted a length-$N$ 1-D signal $[\1_{N/2}^\top,-\1_{N/2}^\top]$ with iid noise, then
solved the optimization $\max_{x_i\in\{-1,1\}}\x^\top \P \x$,
where $\P = [1 \; \c^\top;\c \; \W]$.
Here, $\c$ denotes the noisy 1-D signal, and $\W$ denotes the adjacency matrix corresponding to an unweighted line graph.
We optimized the above objective using SPEC, SDR primal formulation like \eqref{eq:primal}, our proposed GDPA, and SNS.
Results were averaged over $100$ runs.
Table\;\ref{tab:restore} shows that SPEC performed by far the worst at all problem sizes ($N=100$ or $200$), types of line graphs ($1$-hop or $2$-hop neighbor) and noise standard deviation $\sigma$. 
In contrast, GDPA performed similarly to SDR primal and non-SDR scheme SNS.
We thus remove SPEC from experimental comparisons in the sequel.

\begin{figure}[]
\begin{center}
\ifpdf
\includegraphics[width=0.51\linewidth]{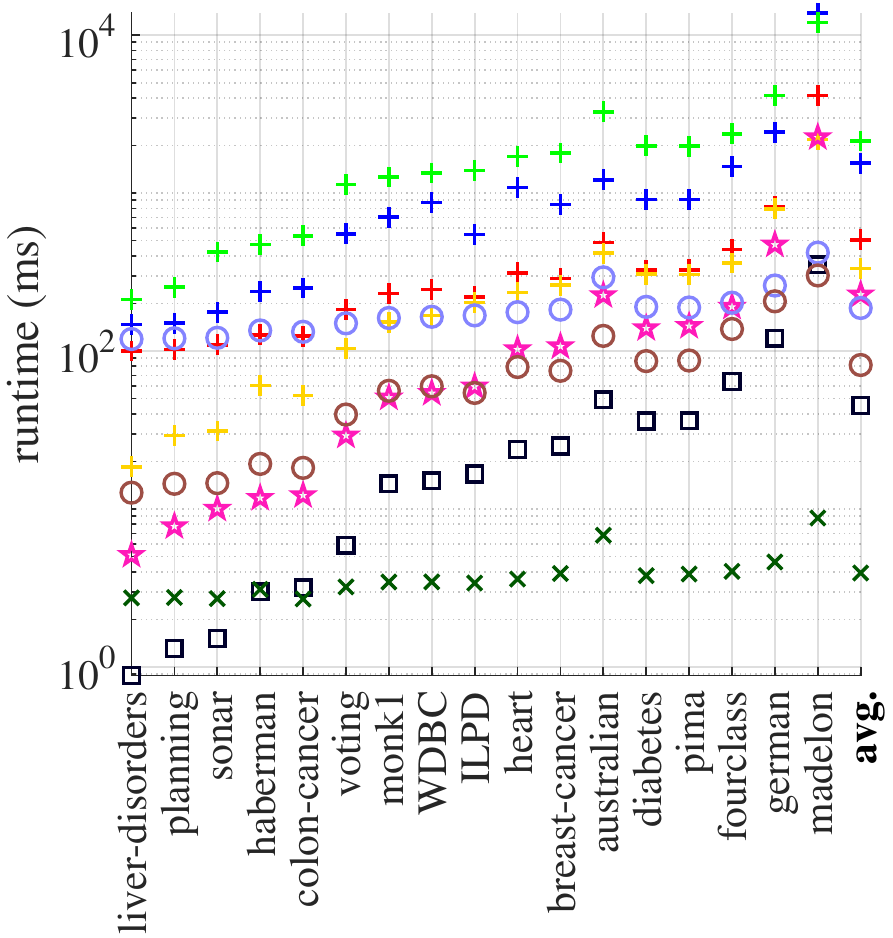}
\includegraphics[width=0.48\linewidth]{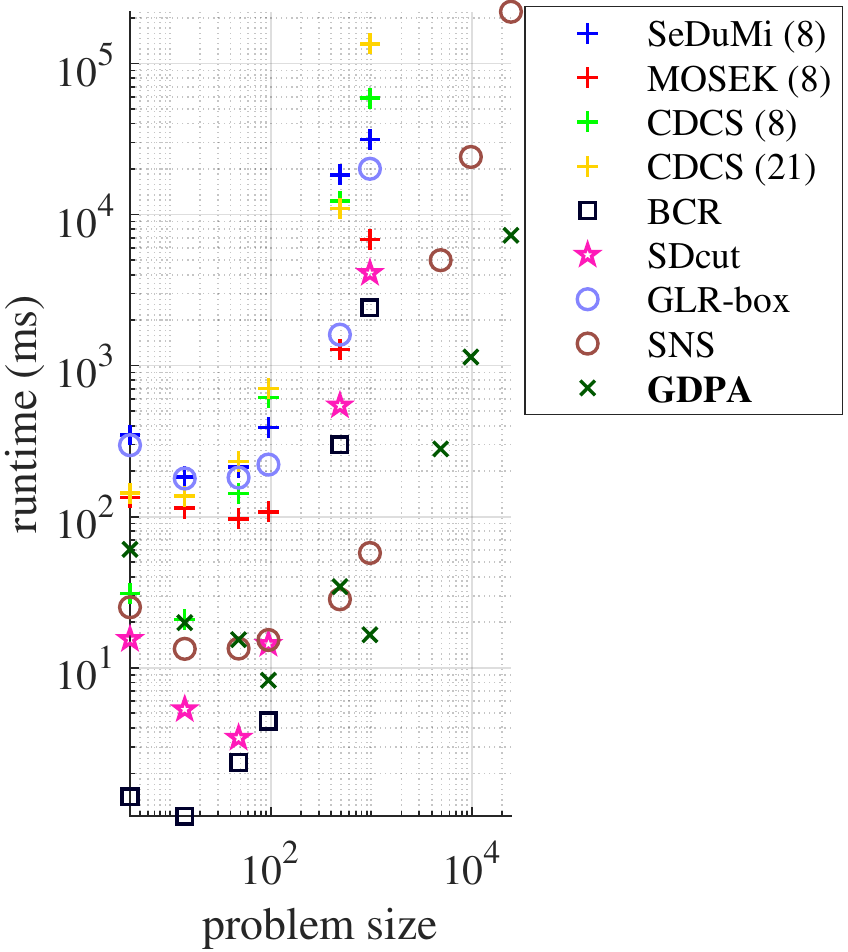}
\else
 do something for regular latex or pdflatex in dvi mode
\fi
\end{center}
\vspace{-0.15in}
\caption{\small Runtime (ms) on 17 datasets (left) with problem sizes $29$ to $400$ and 
\texttt{cod-rna} (right) with problem sizes $4$ to $24428$.}
\label{fig:time}
\end{figure}

We next evaluate competing schemes on classification error and runtime on real datasets.
For each dataset, we first performed \textit{min-max} \cite{10.5555/1671238} and \textit{standardization} \cite{classificationpami19}, two different data re-scaling schemes, to the features of dataset samples, used to compute graph edge weights via an exponential kernel \cite{ortega18ieee}.
For experimental efficiency, we performed a $K$-fold ($K\leq5$) split for each dataset with random seed 0, and then created 10 instances of 50\% training-50\% test split for each fold, with random seeds 1-10 \cite{10.5555/1671238}.
We used 50\% training-50\% test split for each experiment.
See the supplementary file for detailed experimental settings.
Table\;\ref{tab:error_rates} and Fig.\;\ref{fig:time}\,(left) show average classification error rates and runtime (in log scale) of 17 binary datasets \cite{uci_link,libsvm_link} with problem sizes from $29$ to $400$, respectively. 
The $x$-axis of each plot denotes the datasets in ascending order of problem sizes.
Fig.\;\ref{fig:time}\,(right) shows runtime using the same dataset \texttt{cod-rna} with problem sizes from $4$ to $24428$.
We did not execute SeDuMi \eqref{eq:primal}, MOSEK \eqref{eq:primal}, CDCS \eqref{eq:primal}, CDCS \eqref{eq:SDP_dual2}, BCR, SDcut or GLR-box when problem size exceeded $976$.

In terms of classification error rate,
CDCS solving the modified dual \eqref{eq:SDP_dual2} had similar performance as the original SDR primal (SeDuMi \eqref{eq:primal}, MOSEK \eqref{eq:primal} and CDCS \eqref{eq:primal}), showing the validity of our proposed modified SDR dual \eqref{eq:SDP_dual2}.
Further, our proposed GDPA closely approximated the modified SDR dual (CDCS \eqref{eq:SDP_dual2}) in performance, demonstrating the effectiveness of our projection-free GDPA linearization scheme.
By factorizing a PSD matrix $\M = \X \X^{\top}$, BCR avoided tuning of any forward progress step size after each PSD cone projection, which may explain its slightly better average performance.
However, BCR solved a non-convex optimization problem converging to a local minimum, and thus occasionally the performance was relatively poor (\eg, see \texttt{colon-cancer} in the error rate plots in the supplementary file). 
Overall, all solvers performed similarly given constructed similarity graphs in the two cases.

\begin{table}[]
\begin{center}
\caption{Binary signal restoration error (\%).
Original signal $[\1_{N/2}^\top,-\1_{N/2}^\top]$ is 
firstly corrupted using white noise with std $\sigma$ and 
then restored using spectral method SPEC,
SDP solver SeDuMi on SDR primal in \eqref{eq:primal},
proposed GDPA
and
non-SDR method SNS.
Results are averaged over 100 runs.}
\vspace{-0.1in}
\label{tab:restore}
\begin{scriptsize}
\begin{tabular}{c|c|ccc|ccc}
\hline
\multirow{2}{*}{$N$} & graph      & \multicolumn{3}{c|}{1-hop neighbor}                             & \multicolumn{3}{c}{2-hop neighbor}                                                   \\ \cline{2-8} 
                     & $\sigma$   & \multicolumn{1}{c|}{1}     & \multicolumn{1}{c|}{1.5}   & 2     & \multicolumn{1}{c|}{1}     & \multicolumn{1}{c|}{1.5}   & 2                          \\ \hline
\multirow{4}{*}{100} & SPEC        & \multicolumn{1}{c|}{13.04} & \multicolumn{1}{c|}{23.24} & 29.59 & \multicolumn{1}{c|}{10.02} & \multicolumn{1}{c|}{20.94} & 28.06                      \\ \cline{2-8} 
                     & SDR primal & \multicolumn{1}{c|}{2.45}  & \multicolumn{1}{c|}{11.96} & 20.82 & \multicolumn{1}{c|}{0.90}  & \multicolumn{1}{c|}{4.82}  & 12.92                      \\ \cline{2-8} 
                     & \textbf{GDPA}       & \multicolumn{1}{c|}{2.78}  & \multicolumn{1}{c|}{11.01} & 19.57 & \multicolumn{1}{c|}{0.86}  & \multicolumn{1}{c|}{3.57}  & 9.13                       \\ \cline{2-8} 
                     & SNS        & \multicolumn{1}{c|}{1.97}  & \multicolumn{1}{c|}{11.01} & 19.88 & \multicolumn{1}{c|}{0.63}  & \multicolumn{1}{c|}{2.50}  & 8.32                       \\ \hline
\multirow{4}{*}{200} & SPEC       & \multicolumn{1}{c|}{13.54} & \multicolumn{1}{c|}{23.38} & 29.36 & \multicolumn{1}{c|}{11.19} & \multicolumn{1}{c|}{21.69} & 28.20                      \\ \cline{2-8} 
                     & SDR primal & \multicolumn{1}{c|}{1.76}  & \multicolumn{1}{c|}{10.78} & 19.16 & \multicolumn{1}{c|}{0.28}  & \multicolumn{1}{c|}{3.01}  & 10.87 \\ \cline{2-8} 
                     & \textbf{GDPA}       & \multicolumn{1}{c|}{2.46}  & \multicolumn{1}{c|}{10.81} & 18.67 & \multicolumn{1}{c|}{0.39}  & \multicolumn{1}{c|}{2.83}  & 8.52                       \\ \cline{2-8} 
                     & SNS        & \multicolumn{1}{c|}{1.72}  & \multicolumn{1}{c|}{10.92} & 19.09 & \multicolumn{1}{c|}{0.20}  & \multicolumn{1}{c|}{1.48}  & 7.36                       \\ \hline
\end{tabular}
\end{scriptsize}
\end{center}
\end{table}

\begin{table}[]
\begin{center}
\caption{Mean classification error (\%) of 17 binary datasets.}
\vspace{-0.1in}
\label{tab:error_rates}
\begin{tabular}{c|c|c}
\hline
data re-scaling & min-max & standardization \\ \hline
SeDuMi \eqref{eq:primal} & 30.19 & 32.60 \\ \hline
MOSEK \eqref{eq:primal} & 30.31 & 32.61 \\ \hline
CDCS \eqref{eq:primal} & 30.76 & 31.76 \\ \hline
CDCS \eqref{eq:SDP_dual2} & 30.08 & 29.40 \\ \hline
BCR & 27.60 & 26.24 \\ \hline
SDcut & 27.49 & 26.81 \\ \hline
GLR-box & 28.63 & 28.38 \\ \hline
SNS & 33.75 & 30.50 \\ \hline
\textbf{GDPA} & 28.21 & 26.94 \\ \hline
\end{tabular}
\end{center}
\end{table}


In terms of runtime, BCR was competitive with GDPA when the problem size was small, 
but \textit{GDPA significantly outperformed all competing solvers when the problem size was large.}
Specifically, the speed gain increased as problem size increased; for \texttt{madelon} with size $400$, the speedup of GDPA over the next fastest scheme SNS was $34\times$.

Fig.\;\ref{fig:time}\,(right) shows that the computation time for GDPA increased gracefully as the problem size increased to very large sizes.
One reason for our dramatic speed gain is the fast computation of first eigenvectors using LOBPCG, which benefited from warm start.
In general, GDPA performed fewer than 10 LP's until convergence.
In contrast, both CDCS and SDCut required full matrix eigen-decomposition of a matrix of size $N \times N$ per iteration; 
the speedup of replacing the full eigen-decomposition with one LP plus first eigenvector computation per iteration was significant. 
For BCR, each iteration required either $N$-dimensional matrix inversion for a least-squares problem or iterative gradient descent, which was computationally expensive as the problem size increased.
SNS required many matrix-vector multiplications---which was time-consuming as the problem size increased---though manually adjusting the number of neighborhood vectors can potentially improve speed. 
On average, GDPA enjoyed a $28\times$ speedup over the next fastest solver SNS.

On average, the difference between $\lambda_{\min}(\H)$ and $\lambda_{\min}(\bar{\H})$ in Lemma \ref{lemma:Hbar} is $1.1608\times 10^{-7}$, which is very small. This demonstrates the tightness of bound $\lambda_{\min}(\bar{\H}) \leq \lambda_{\min}(\H)$ in practice, and thus the effectiveness of Lemma\;\ref{lemma:Hbar}.

\section{Conclusion}
\label{sec:conclude}
We propose a fast projection-free algorithm for the binary graph classification problem.
The key idea is to replace the difficult positive semi-definite (PSD) cone constraint with linear constraints derived from the recent Gershgorin disc perfect alignment (GDPA) theory, so that each iteration requires only one linear program (LP) and one first eigenvector computation. 
Experiments show that our algorithm enjoyed on average $28 \times$ speedup over the next fastest competitor while retaining comparable label prediction performance.

As an optimization problem, binary graph classification is rather narrowly defined 
(though multi-class classification can be implemented as a tree of binary classifiers).
Further, performance depends heavily on the construction of a good similarity graph, which is outside this paper's scope. 
However, we conjecture that the general methodology of GDPA linearization can be similarly tailored to other QCQP problems with PSD cone constraints.
We anticipate that speedups in other QCQP problems will also be significant.


\bibliography{aaai22}


\end{document}